\documentclass[letterpaper, 10 pt, conference]{ieeeconf}  

\IEEEoverridecommandlockouts 

\usepackage{graphics} 
\usepackage{epsfig} 
\usepackage{mathptmx} 
\usepackage{times} 
\usepackage{amsmath} 
\usepackage{amssymb}  
\usepackage{multirow}
\usepackage{subcaption}

\title{\LARGE \bf
Toward Signing Activity Projection in Sign Language Interaction
}

\author{Takao Obi$^{1}$, Wang Yusong$^{1}$, Koji Inoue$^{2}$ and Kotaro Funakoshi$^{1}$ 
\thanks{$^{1}$Takao Obi, Wang Yusong, and Kotaro Funakoshi are with Institute of Science Tokyo, Yokohama 230-0045, Japan.
        {\tt\small \{smalltail, wangyi, funakoshi\}@lr.first.iir.isct.ac.jp}}%
\thanks{$^{2}$Koji Inoue is with Kyoto University, Yoshida-honmachi, Sakyoku, Kyoto, Japan.
        {\tt\small inoue.koji.3x@kyoto-u.ac.jp}}%
}

\begin{document}

\maketitle
\thispagestyle{empty}
\pagestyle{empty}

\begin{abstract}
Social robots must interact robustly not only with users assumed by speech-centered systems but also with diverse users whose communication relies on different modalities, e.g., sign language. 
One important capability gap is predictive turn-taking with signing users.
Although Voice Activity Projection (VAP) has been successfully used to model future voice activity in spoken interaction, it remains unclear whether the framework transfers to sign language interaction. 
This paper presents an initial transfer study of adapting a VAP architecture to dyadic sign language interaction.
Using interaction recordings from the Public DGS Corpus, we derive binary signing activity streams from lexical sign annotations and formulate proxy tasks for turn-taking prediction. 
The model uses pose-derived hand, eye-region, and mouth-region features extracted for each signer. 
The results show that SHIFT/HOLD prediction is promising, especially with hand cues, while SHIFT-prediction remains difficult. 
These findings provide initial evidence for both the promise and the current limitations of transferring predictive turn-taking models from spoken interaction to sign language interaction.
Predictive modeling of sign language interaction still requires sign-language-specific event definitions that go beyond speech-derived categories.
\end{abstract}

\section{Introduction}
Toward human–robot symbiosis, social robots must interact robustly not only with the users implicitly assumed by current systems, but also with diverse users whose communicative modalities differ from the speech-centered assumptions of existing interaction pipelines. 
Recent work in spoken interaction has proposed Voice Activity Projection (VAP), a predictive framework that models the future voice activity of both interlocutors and supports turn-taking tasks such as SHIFT/HOLD and SHIFT-prediction~\cite{Ekstedt22_interspeech}. 
Concretely, SHIFT/HOLD asks whether, after a period of mutual inactivity, the same speaker as before the silence resumes (HOLD) or the turn passes to the other speaker (SHIFT), whereas SHIFT-prediction asks whether an ongoing activity is likely to be followed by a turn shift in the near future.
VAP models have been investigated for real-time turn-taking control in spoken dialogue systems and dialogue robots~\cite{Inoue24_realtime, Skantze25}, highlighting the value of predictive timing control for smoother interaction.

However, turn-taking is not unique to speech. 
Prior studies have shown that sign language interactions are also organized through systematic turn-taking, with modality-specific cues such as hand movement phases, turn-final holds, and eye gaze~\cite{Coates01, Groeber14}. 
Moreover, when sign-language-specific movement phases are taken into account, turn timing in sign language question–answer sequences can closely resemble that of spoken interaction~\cite{deVos15}. 
Despite this, activity projection for sign language interaction remains largely unexplored.

In this paper, we take a first step toward adapting the activity projection framework to sign language interaction. 
Using sign language interaction data from the Public DGS Corpus~\cite{DGSCorpus}, we formulate proxy turn-taking tasks from signing and non-signing states. 
We represent each interlocutor using pose-derived hand, eye-region, and mouth-region cues, and evaluate whether these visual signals support predictive turn-taking in sign language interaction.
Our experiments show that SHIFT/HOLD 
on sign language interaction proxies is promising, especially when hand cues are used, whereas SHIFT-prediction remains challenging. 
These findings provide initial evidence for both the promise and the current limits of transferring predictive turn-taking models from spoken interaction to sign language interaction.
Predictive modeling of sign language interaction still requires sign-language-specific event definitions that go beyond speech-derived categories.

\section{Related Work}

\subsection{Turn-taking in sign language interaction}
Research on sign language interaction has shown that sign language interactions are not unstructured, but are organized through systematic timing, floor management, and participant coordination.
Early studies of deaf interaction documented regularities in overlap, floor management, and turn exchange, arguing that signers also orient to orderly turn exchange rather than unconstrained simultaneous production~\cite{Coates01}.
More recent corpus work likewise suggests that sign language interaction is broadly compatible with one-at-a-time turn-taking organization, although boundary identification is often complicated by non-semantic movement~\cite{Green23}.

Within this broader interactional organization, prior work has identified several modality-specific resources for turn-taking.
Interaction analytic and phonetic studies describe turn-final holds, preparation phases, overlap resolution near possible completion points, and gaze behavior as important resources for coordinating turn transitions~\cite{Groeber14, deVos15, Girard15, Kikuchi08}.
In particular, de~Vos and Levinson showed that when stroke-to-stroke boundaries are considered, turn timing in sign language question--answer sequences can closely resemble the tight timing observed in spoken interaction~\cite{deVos15}.
Experimental work further showed that both signers and non-signers can anticipate turn ends in sign language interaction~\cite{deVos22}.

The inventory of relevant interactional cues is broader than turn-final holds alone.
Ferrara showed that finger-pointing in sign language interaction can serve several interactional functions, including giving the turn, taking the turn, keeping a turn open, and holding the turn~\cite{Ferrara20}.
Ferrara also described how finger-pointing can index turn beginnings in Norwegian Sign Language interaction~\cite{Ferrara22}.
In addition, Arnold and Ferrara reported that finger-pointing and palm-up actions frequently occur in question contexts, where they can help manage responses and interactional expectations~\cite{Arnold24}.

Research on feedback behavior has likewise shown that sign language interaction includes dedicated listener-side signals.
Mesch analyzed manual backchannel responses in Swedish Sign Language~\cite{Mesch16}, Borstell proposed corpus methods for identifying continuers in Swedish Sign Language~\cite{Borstell24}, and Bauer et al. showed that feedback head nods and affirmative nods in German Sign Language (DGS) differ phonetically and can be studied using pose estimation~\cite{Bauer24}.
These findings on sign language interaction show that it contains rich manual and non-manual resources for regulating turn-taking.

\subsection{Signing activity segmentation}
Work on sign language processing has also addressed how activity and larger units should be segmented.
On the engineering side, sign language detection aims to determine whether signing is present in each frame, and pose-based methods have enabled real-time detection from body-joint motion~\cite{Moryossef20}.
Subsequent work argues that continuous signing is not adequately captured by simple inside/outside labeling alone and that linguistically motivated sign- and phrase-level segmentation is needed~\cite{Moryossef23}.
On the corpus and interaction side, Bono et al. proposed multimodal utterance-unit annotation for Japanese Sign Language interaction, using signing, mouth movements, non-manual movements, and gaze to identify interactional boundaries rather than relying on spoken-language translations~\cite{Bono20}.
Related corpus work likewise shows that lexical sign annotations do not trivially yield higher-level utterance units~\cite{Borstell24_align}.
Taken together, this literature suggests that sign language interaction provides systematic cues for turn management, but these cues have not yet been operationalized as a standardized computational target for continuous turn-taking prediction.

\subsection{Voice Activity Projection for predictive turn-taking}
Voice Activity Projection (VAP) formulates turn-taking as the continuous prediction of the future voice activity of both interlocutors~\cite{Ekstedt22_interspeech}.
Rather than predicting a single end-of-turn point, VAP predicts short-horizon future activity patterns and supports downstream tasks such as SHIFT/HOLD, SHIFT-prediction, SHORT/LONG, and Backchannel-prediction.
Because the model is trained from future activity itself, it avoids the need for manually annotated turn-taking labels and has become a strong framework for spoken turn-taking modeling.

Subsequent work has extended VAP in several directions.
These include multilingual training across English, Mandarin, and Japanese~\cite{Inoue24_multilingual}, real-time continuous inference for dialogue systems~\cite{Inoue24_realtime}, and multimodal spoken settings that incorporate non-verbal cues such as visual cues and biosignals~\cite{Onishi23,Obi25}.
VAP-style models have also begun moving from offline evaluation to interactive use.
General turn-taking models, including VAP, have improved response timing and reduced interruptions in human--robot interaction without task-specific fine-tuning~\cite{Skantze25}, and noise-robust VAP-based systems have improved response latency in real-world dialogue robot field experiments~\cite{Inoue25}.

Despite this progress, current VAP research remains centered on speech and future voice activity.
Even multimodal extensions are still developed and evaluated in spoken settings, where verbal activity provides the primary target and supervisory signal.
It therefore remains unclear how the VAP formulation should be transferred to sign language interaction, where turn boundaries are expressed visually, relevant cues are distributed across manual and non-manual channels, and sign-language-specific interactional events have not yet been standardized in a VAP-style evaluation framework.

\section{Experiment}
This study provides an initial test of whether a turn-taking model can be transferred from spoken interaction to sign language interaction.
We replaced audio with pose-based visual cues extracted from video and modeled future signing activity in the same formulation as Voice Activity Projection (VAP).

\subsection{Dataset}
\begin{figure}[t]
  \includegraphics[width=\linewidth]{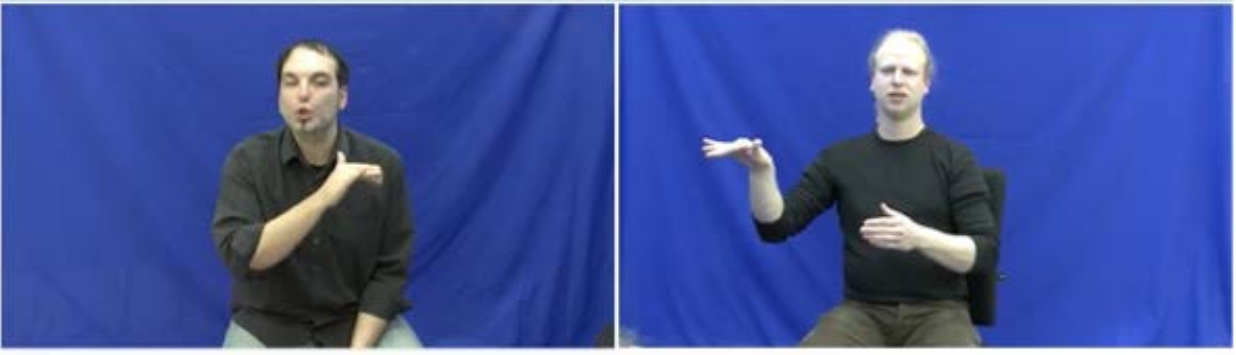}
  \caption{Example of sign language interaction in Public DGS Corpus.}
  \label{fig:dial_example}
\end{figure}

We used the dialogue-tagged subset of the Public DGS Corpus~\cite{DGSCorpus}, which contains spontaneous dyadic sign language interaction.
For each recording, the corpus provides OpenPose-based 2D keypoints for both signers, including hand and facial regions, together with manual annotation tiers that mark temporal segments of lexical signing activity.
Figure~\ref{fig:dial_example} shows an example interaction from the Public DGS Corpus.

Following prior work on pose-based signing activity detection~\cite{Moryossef20}, we constructed a binary signing activity stream for each signer by taking the union of all annotated sign segments.
A frame was labeled as active if its timestamp fell within any annotated segment, and inactive otherwise.
We used 2D hand keypoints and selected facial keypoints corresponding to the eye and mouth regions, normalized in image coordinates.
Missing values were replaced with zeros.
We resampled labels to 30\,fps to match the existing VAP multimodal model frame rate~\cite{Obi25}.

We first split the data at the transcript level into train, validation, and test sets (80\%/10\%/10\%), and then extracted fixed-length 20\,s windows (stride = 20\,s) from each split, following the existing VAP preprocessing~\cite{Inoue24_realtime}.
Table~\ref{tab:event_stats} reports dataset statistics and the numbers of extracted event targets (not the number of 20\,s windows) computed on the resulting splits.
For SHIFT-prediction, negative samples are constructed to match the number of positives; we report only positive counts.

\begin{table}[t]
\centering
\caption{Dataset statistics and event target counts. Event targets were derived from the binary activity streams following the standard VAP evaluation protocol.}
\label{tab:event_stats}
\begin{tabular}{lcccc}
\hline
\multirow{2}{*}{Dataset} & Number of & Duration & SHIFT /  & SHIFT- \\
& dialogues & [h] & HOLD & prediction\\
\hline
Train & 199 & 41.23 & 20,333 / 61,749 & 5,010 \\
Val & 25 & 4.53 & 2,080 / 6,081 & 540 \\
Test & 23 & 3.89 & 2,111 / 6,717 & 780 \\
\hline
\end{tabular}
\end{table}

Importantly, these labels represent signing activity rather than utterance-unit or turn-unit annotations.
Prior work on sign language interaction has shown that interactional boundaries are multimodal and are not trivially recoverable from lexical sign segmentation alone~\cite{Bono20}.

\subsection{Models}
\begin{figure}[t]
    \centering
    \includegraphics[width=\linewidth]{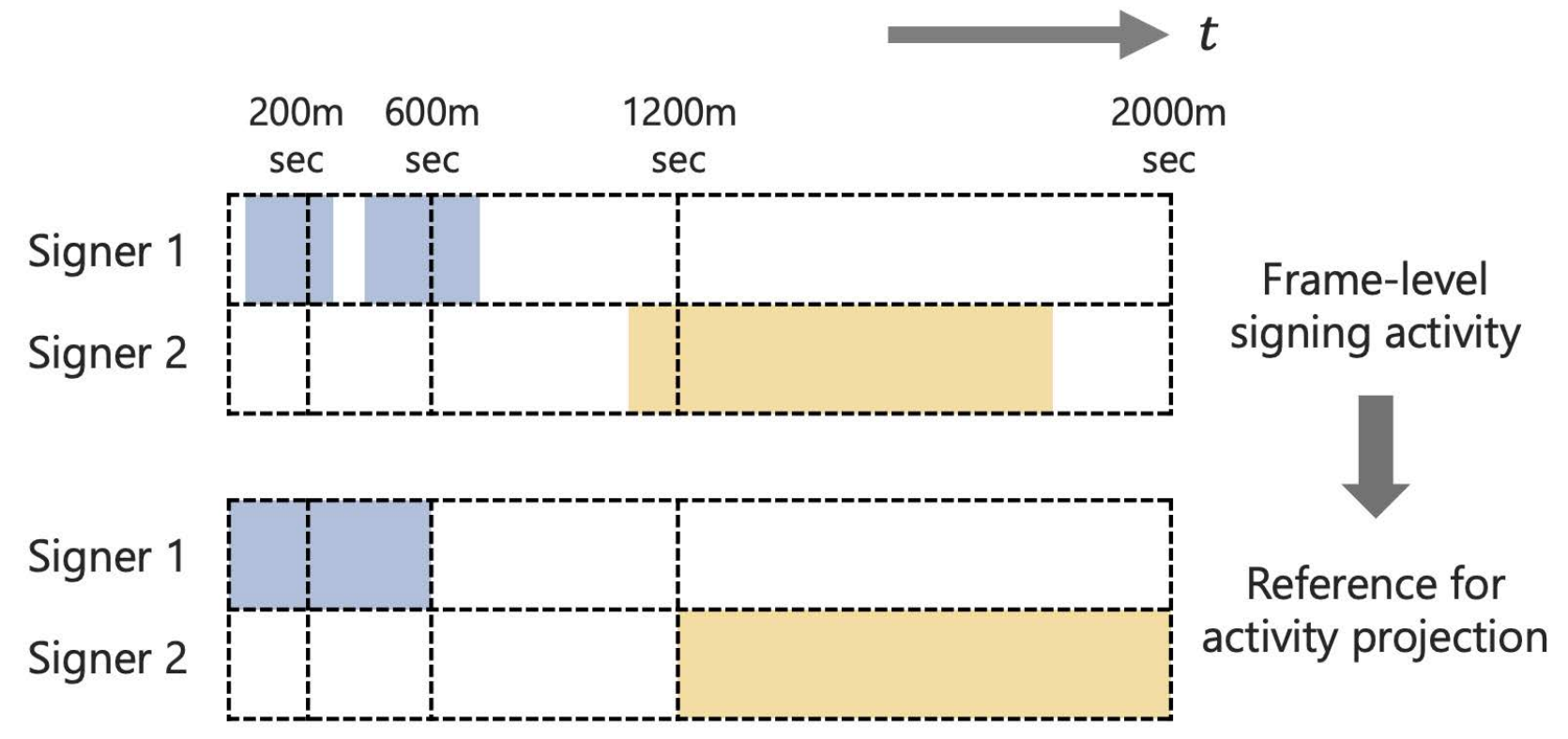}
    \caption{Projection binning scheme.}
    \label{fig:window_binary}
\end{figure}

We adopt the original VAP formulation for dyadic sign language interaction:
Given a context window, the model predicts a discrete future activity state over a 2\,s horizon, discretized into four future bins with durations of 0.2, 0.4, 0.6, and 0.8\,s (cumulative boundaries at 0.2, 0.6, 1.2, and 2.0\,s).
Each bin was labeled as signing if at least 50\% of the frames in that bin were labeled as signing, and as non-signing otherwise, 
following the original VAP discretization~\cite{Ekstedt22_interspeech}, as depicted in Figure~\ref{fig:window_binary}.
This yields a 256-class state space (8 binary decisions: 2 signers $\times$ 4 bins).

\begin{figure}[t]
    \centering
    \includegraphics[width=\linewidth]{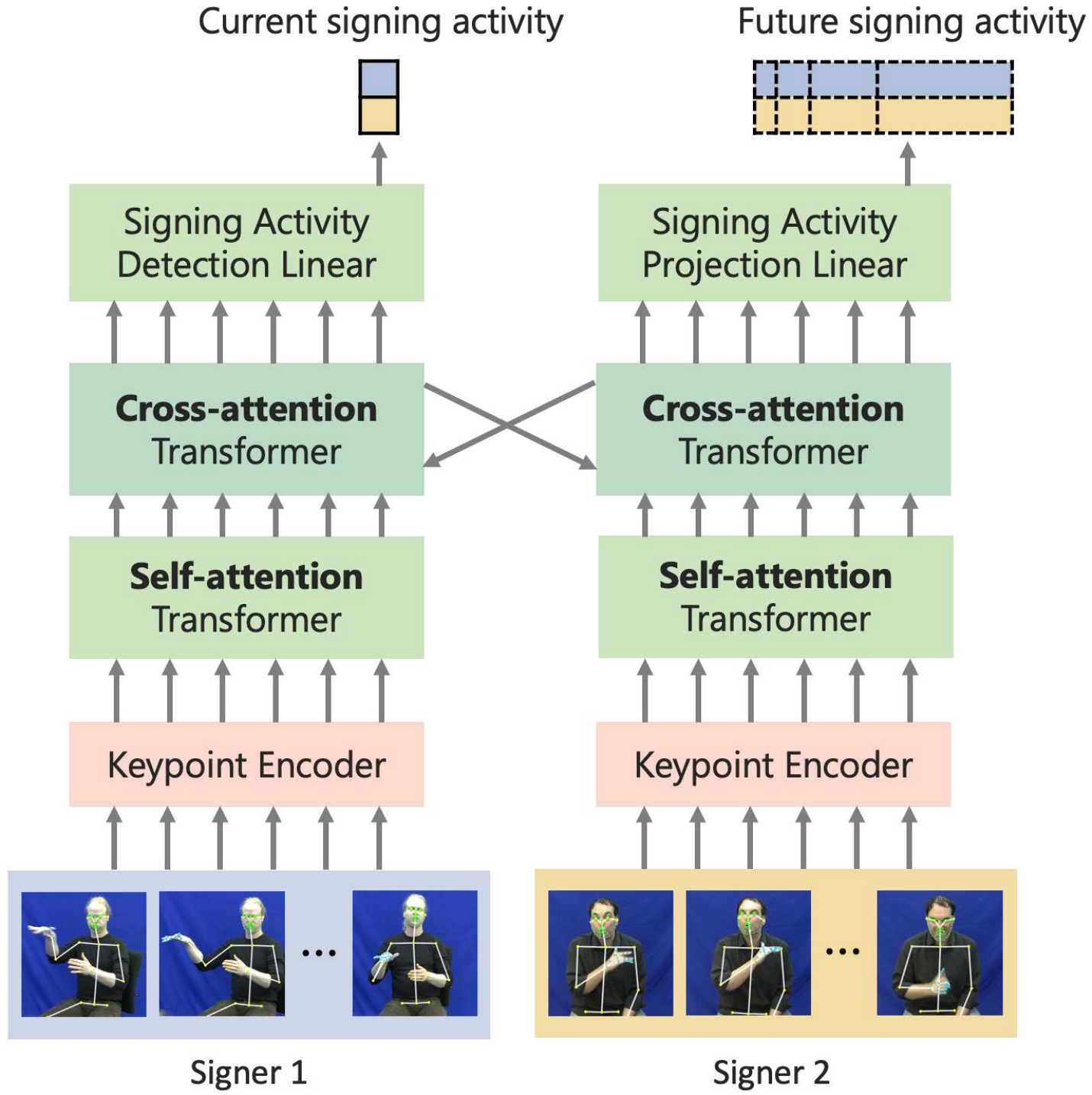}
    \caption{Model architecture overview.}
    \label{fig:model_overview}
\end{figure}

We started from an existing VAP implementation\footnote{https://github.com/MaAI-Kyoto/MaAI/tree/main/train} and replaced the audio encoder with a pose encoder.
Figure~\ref{fig:model_overview} illustrates the model architecture. 
The model consists of four main components:

\noindent\textbf{Keypoint Encoder}:
A modality-specific linear projection (LayerNorm + Linear + GELU + Dropout) to map raw features to a 256-dimensional embedding. 

\noindent\textbf{Self-attention Transformer}: 
A single custom self-attention Transformer layer with 256 dimensions to model each signer's pose stream separately.

\noindent\textbf{Cross-attention Transformer}: 
Three custom Transformer layers with 256 dimensions that perform cross-attention over both signers' encoded pose streams.

\noindent\textbf{Prediction heads}: 
Two linear output heads were used. One for the 256-class future activity state $p_{\mathrm{proj}}(y)$ and one for per-signer activity detection $p_{\mathrm{signing}}(s)$.

The model losses are defined as $L = L_{\mathrm{proj}} + L_{\mathrm{signing}}$, where
\begin{align}
L_{\mathrm{proj}} = &-\log p_{\mathrm{proj}}(y), \notag\\
L_{\mathrm{signing}} = &- \sum_{s=1}^{2} \Bigl[v_s \log p_{\mathrm{signing}}(s) \notag\\
&+ (1 - v_s)\log\bigl(1 - p_{\mathrm{signing}}(s)\bigr)\Bigr], \notag
\end{align}
$y \in \{1,\dots,256\}$ is the reference future activity state, and $v_s \in \{0,1\}$ indicates whether signer $s$ is active. 
For brevity, the time frame indexing is omitted, but these calculations apply to all input frames.

We evaluated multiple modality settings by selecting different subsets of the pose input: hands, eye-region keypoints, mouth-region keypoints, and their combinations.
Following the existing multimodal VAP framework~\cite{Obi25}, each modality was processed with its own encoder and attention stack, and the resulting cross-attended representations were fused by concatenation before prediction.

\subsection{Procedure}
We trained with 20\,s windows (batch size 8) using AdamW (learning rate $3.63\times 10^{-4}$, weight decay $10^{-3}$).
We trained each model for up to 30 epochs and selected the checkpoint with the lowest validation loss for testing.
The maximum number of epochs was chosen based on preliminary experiments, which showed that the validation loss decreased in the early stage of training and later increased in some runs due to overfitting.
To estimate training stability, we repeated training with seeds 1--10 and obtained the mean and standard deviation across runs.

\subsection{Evaluation}
We evaluated turn-taking performance following the standard VAP protocol, deriving event labels from the binary signing activity streams.
In this initial study, we focused on two tasks: SHIFT/HOLD and SHIFT-prediction.
SHIFT/HOLD evaluates whether the next activity after a region of mutual inactivity is a HOLD by the same signer as before the inactivity or a SHIFT to the interlocutor, while 
SHIFT-prediction evaluates whether an ongoing activity is likely to be followed by a turn shift within the next 600\,ms.
Because the SHIFT/HOLD event distributions are imbalanced, we report balanced accuracy and macro-F1 as the primary metrics, along with the test cross-entropy loss.

We did not evaluate SHORT/LONG or Backchannel-prediction in this study.
Under the present signing/non-signing proxy, SHORT/LONG would reduce to the task of distinguishing shorter and longer stretches of lexical signing activity, whereas prior work shows that interactional boundaries and higher-level utterance units in sign language are multimodal~\cite{Bono20} and are not trivially recoverable from lexical sign segmentation alone~\cite{Borstell24_align}.
In addition, prior work has documented sign-language-specific backchannel responses in sign language interaction~\cite{Mesch16}, but the current corpus setting provides only binary signing activity streams derived from lexical sign annotations.
These streams do not specify which active segments function as backchannels.
Therefore, evaluating these tasks would require dedicated event definitions beyond the speech-derived VAP categories.

\subsection{Results}
Table~\ref{tab:result} summarizes the model performance of each modality configuration.
The constant baseline always predicts HOLD in SHIFT/HOLD and no-shift in SHIFT-prediction.
For SHIFT/HOLD, only models including hand cues improved consistently over the constant baseline on both balanced accuracy and macro-F1. 
Performance on SHIFT/HOLD was strongest when hand cues were included. 
For SHIFT-prediction, all learned models outperformed the constant baseline on both metrics. 
The eye-only configuration achieved the highest mean scores, although the margin over the other configurations was small.

\begin{table*}[t]
    \centering
    \caption{Test loss, balanced accuracy, and macro-F1 for each modality configuration (mean $\pm$ standard deviation over 10 seeds).}
    \label{tab:result}
    \small
    \begin{tabular}{c|ccc|c|cc|cc}
        \hline
        \multirow{2}{*}{Model} & \multicolumn{3}{c|}{Features} & \multirow{2}{*}{Test CE Loss} & \multicolumn{2}{c|}{SHIFT / HOLD} & \multicolumn{2}{c}{SHIFT-pred} \\
        \cline{2-4}
        \cline{6-9}
        & Hands & Eyes & Mouth &  & bal-acc & macro-F1 & bal-acc & macro-F1 \\
        \hline
        Constant baseline & -- & -- & -- & -- & 0.500 & 0.424 & 0.500 & 0.333 \\
        \hline
        \multirow{7}{*}{Proposed} & $\circ$ & -- & -- & 3.019 $\pm$ 0.010 & \textbf{0.646} $\pm$ 0.018 & \textbf{0.501} $\pm$ 0.015 & 0.552 $\pm$ 0.012 & 0.453 $\pm$ 0.023 \\
        & -- & $\circ$ & -- & 3.378 $\pm$ 0.008 & 0.520 $\pm$ 0.027 & 0.380 $\pm$ 0.016 & \textbf{0.558} $\pm$ 0.015 & \textbf{0.493} $\pm$ 0.022 \\
        & -- & -- & $\circ$ & 3.215 $\pm$ 0.009 & 0.546 $\pm$ 0.020 & 0.380 $\pm$ 0.011 & 0.540 $\pm$ 0.012 & 0.433 $\pm$ 0.018 \\
        & $\circ$ & $\circ$ & -- & 3.013 $\pm$ 0.010 & 0.640 $\pm$ 0.013 & 0.492 $\pm$ 0.016 & 0.549 $\pm$ 0.012 & 0.448 $\pm$ 0.014 \\
        & $\circ$ & -- & $\circ$ & 3.004 $\pm$ 0.006 & 0.640 $\pm$ 0.019 & 0.496 $\pm$ 0.019 & 0.545 $\pm$ 0.010 & 0.443 $\pm$ 0.015 \\
        & -- & $\circ$ & $\circ$ & 3.209 $\pm$ 0.006 & 0.554 $\pm$ 0.018 & 0.389 $\pm$ 0.012 & 0.543 $\pm$ 0.013 & 0.436 $\pm$ 0.020 \\
        & $\circ$ & $\circ$ & $\circ$ & 3.007 $\pm$ 0.010 & 0.644 $\pm$ 0.019 & \textbf{0.501} $\pm$ 0.020 & 0.539 $\pm$ 0.011 & 0.432 $\pm$ 0.017 \\
        \hline
    \end{tabular}
\end{table*}

\section{Discussion}
The results suggest that transferring a VAP-style predictive framework to sign language interaction is promising, but the success of the transfer depends strongly on how the target is defined.
Hand-only input achieved the highest mean score on SHIFT/HOLD, indicating that pose-derived manual cues are informative for estimating the current local turn state under the present proxy labels.
At the same time, this result should be interpreted with care: because the binary targets were constructed from lexical sign annotations, the task is naturally more aligned with manual activity than with non-manual behaviors.

By contrast, SHIFT-prediction remained substantially more difficult.
The eye-only model achieved the highest mean score on SHIFT-prediction, but the margin over the other conditions was small. 
This suggests that anticipatory information may be present in non-manual cues; however, this does not imply that manual information is uninformative. 
The current proxy labels, pose-based hand representations, limited data, and model capacity may be insufficient to reliably exploit manual structure.

Combining additional modalities with hands did not consistently improve performance.
In the current setting, this likely reflects a combination of factors: noisier non-manual keypoints, limited training data, and a mismatch between signing activity-based proxy labels and granular interactional events.
Overall, the results support the use of signing activity as an initial proxy for transferring spoken turn-taking models, while also underscoring the need for sign-language-specific definitions and richer multimodal representations.

\section{Limitations}
\subsection{Proxy signing labels}
Our labels were proxy labels obtained by taking the union of annotated lexical sign segments and deriving VAP-style events from the resulting binary activity streams, following the general idea of pose-based signing activity detection~\cite{Moryossef20}.
This makes the task reproducible on an existing corpus, but these labels should not be interpreted as utterance units, turn units, or sign-language-specific interactional events.
Moreover, because the targets are anchored in lexical sign annotations, they are naturally closer to manual activity than to non-manual behaviors.
Accordingly, the current proxy labels may overstate the apparent contribution of hand cues and understate the contribution of eye- and mouth-related cues.

\subsection{Speech-derived projection formulation}
In this study, we adopt VAP as an initial step toward predictive turn-taking modeling for sign language interaction, but the original VAP formulation was developed for future voice activity.
Although we replace the audio input with pose-derived visual input, the projection formulation itself remains inherited from spoken-interaction settings.
Specifically, future activity is discretized into fixed-duration projection bins and represented as binary active/inactive states, using temporal boundaries originally introduced for voice activity.
We retained this formulation for comparability with prior work, but it may not provide the most appropriate temporal representation for signing activity, which may unfold over different timescales and interactional units than speech-derived voice activity.

\subsection{Pose-derived model inputs}
Because the input is based on image-normalized 2D keypoints, it cannot directly capture depth, fine-grained handshape, or precise gaze direction.
We used pose-derived hand, eye-region, and mouth-region cues, but they provide only a coarse approximation of the information available in sign language interaction.
Eye-region keypoints are not gaze estimates, and important non-manual behaviors such as head nods, brow actions, and mouthings may require more dedicated representations.

\subsection{Scope of experimental settings}
Our study is an initial transfer study, but it does not establish generality across sign languages, recording conditions, or interactive settings.
We evaluated a single corpus, a dyadic interaction subset, and two proxy tasks inherited from speech-oriented VAP evaluation.
Also, the current transcript-level split may not fully isolate signer-specific generalization.

The motivation of this study is to reduce speech-centered capability gaps in social robots, but the present experiments remain offline and corpus-based.
The reported results do not show that a robot can yet manage sign language turn-taking robustly in live interaction, nor how projected activity should be translated into control decisions such as holding, yielding, claiming, or repairing a turn.

\section{Future Work}
\subsection{Construction of sign-language-specific interaction targets}
A direct next step is to define sign-language-specific interactional events and construct evaluation targets based on these definitions rather than on signing activity alone.
Prior work on sign language interaction has shown that interactional boundaries are multimodal and are not trivially recoverable from lexical sign segmentation alone~\cite{Bono20}.
More generally, analyses of turn organization in sign language interaction suggest that gaze, head movement, and other non-manual behaviors may contribute to how turns are projected, held, and yielded~\cite{Kikuchi08}.
A useful extension of the present framework would therefore be to define richer interactional state labels that better reflect the broader organization of sign language interaction, rather than relying only on the presence or absence of lexical signs.
Once such labels are defined, the projection formulation itself should also be redesigned around them, rather than directly reusing speech-derived projection bins.

\subsection{Expansion of input representation}
Future work should compare keypoint-based models with richer video encoders, explicit head and gaze features, and architectures that model manual and non-manual channels separately.
More expressive video representations may capture interactional cues that are difficult to represent through sparse 2D keypoints alone.
In addition, richer manual representations, such as gloss-level features derived from sign language recognition, may help the model exploit linguistic structure more reliably.
If such signals cannot be extracted reliably from existing recordings, progress may require new corpora with denser multimodal annotation, higher-resolution video, or sensing setups designed to capture fine-grained hand and face behavior more directly.

\subsection{Analysis in multiple settings}
Future work should expand the analysis to additional sign language corpora and evaluate the resulting models across languages, recording conditions, and interactional settings.
The relative usefulness of manual and non-manual cues, as well as the calibration of projected activity, may vary across sign languages and datasets.

\subsection{From offline prediction to embodied robot interaction}
Future work should integrate the predictor into a real-time robot and evaluate it with deaf and hard-of-hearing users in actual interaction, using not only offline prediction metrics but also interaction-level measures across diverse users and recording conditions.
In deployment to reduce capability gaps, a robot must estimate these cues online under viewpoint changes, occlusion, tracking failures, and system latency, while its own embodiment, gaze, and arm motions can also affect the visual channel used by the signer.

\section{Conclusion}
This paper presented an initial transfer study of adapting Voice Activity Projection to sign language interaction. 
Using interaction data from the Public DGS Corpus, we formulated proxy turn-taking prediction tasks from signing activity. 
The results showed that SHIFT/HOLD prediction is promising, especially with hand cues, whereas SHIFT-prediction remains difficult under the current formulation. 
These findings suggest that predictive turn-taking can be transferred to sign language interaction only partially under speech-derived event definitions.
To improve robot interaction with deaf and hard-of-hearing users, future work will require moving beyond proxy signing activities toward sign-language-specific interactional targets.


\section*{ACKNOWLEDGMENT}
This work was supported by Japan Science and Technology Agency (JST) as part of Adopting Sustainable Partnerships for Innovative Research Ecosystem (ASPIRE), Grant Number JPMJAP25B3.
This study was carried out using the TSUBAME4.0 supercomputer at Institute of Science Tokyo.

\bibliographystyle{IEEEtran}
\bibliography{reference}

\end{document}